\documentclass[11pt]{article}
\usepackage{acl2016}
\usepackage{times}
\usepackage{url}
\usepackage{latexsym}
\usepackage{graphicx}
\usepackage{amsmath}
\usepackage{comment}

\aclfinalcopy % Uncomment this line for the final submission
\title{ Joint Online Spoken Language Understanding and Language Modeling with Recurrent Neural Networks}

\author{Bing Liu \\
  Carnegie Mellon University \\
  Electrical and Computer Engineering \\
  {\tt liubing@cmu.edu} \\\And
  Ian Lane \\
  Carnegie Mellon University \\
  Electrical and Computer Engineering \\
  Language Technologies Institute \\
  {\tt lane@cmu.edu} \\}

\date{}

\begin{document}
\maketitle
\begin{abstract}
    Speaker intent detection and semantic slot filling are two critical tasks in spoken language understanding (SLU) for dialogue systems. In this paper, we describe a recurrent neural network (RNN) model that jointly performs intent detection, slot filling, and language modeling. The neural network model keeps updating the intent prediction as word in the transcribed utterance arrives and uses it as contextual features in the joint model. Evaluation of the language model and online SLU model is made on the ATIS benchmarking data set. On language modeling task, our joint model achieves 11.8\% relative reduction on perplexity comparing to the independent training language model. On SLU tasks, our joint model outperforms the independent task training model by 22.3\% on intent detection error rate, with slight degradation on slot filling F1 score. The joint model also shows advantageous performance in the realistic ASR settings with noisy speech input.
\end{abstract}

\section{Introduction}
    As a critical component in spoken dialogue systems, spoken language understanding (SLU) system interprets the semantic meanings conveyed by speech signals. Major components in SLU systems include identifying speaker's intent and extracting semantic constituents from the natural language query, two tasks that are often referred to as intent detection and slot filling. 
    
    Intent detection can be treated as a semantic utterance classification problem, and slot filling can be treated as a sequence labeling task. These two tasks are usually processed separately by different models. For intent detection, a number of standard classifiers can be applied, such as support vector machines (SVMs)~\cite{haffner:03} and convolutional neural networks (CNNs)~\cite{xu:13}. For slot filling, popular approaches include using sequence models such as maximum entropy Markov models (MEMMs)~\cite{mccallum:00}, conditional random fields (CRFs)~\cite{raymond:07}, and recurrent neural networks (RNNs)~\cite{yao:14,mesnil:15}. 
    
    Recently, neural network based models that jointly perform intent detection and slot filling have been reported. Xu~\shortcite{xu:13} proposed using CNN based triangular CRF for joint intent detection and slot filling. Guo~\shortcite{guo:14} proposed using a recursive neural network (RecNN) that learns hierarchical representations of the input text for the joint task. Such joint models simplify SLU systems, as only one model needs to be trained and deployed. 
    
    The previously proposed joint SLU models, however, are unsuitable for \textit{online} tasks where it is desired to produce outputs as the input sequence arrives. In speech recognition, instead of receiving the transcribed text at the end of the speech, users typically prefer to see the ongoing transcription while speaking. In spoken language understanding, with real time intent identification and semantic constituents extraction, the downstream systems will be able to perform corresponding search or query while the user dictates. The joint SLU models proposed in previous work  typically require intent and slot label predictions to be conditioned on the entire transcribed word sequence. This limits the usage of these models in the online setting.

    In this paper, we propose an RNN-based online joint SLU model that performs intent detection and slot filling as the input word arrives. In addition, we suggest that the generated intent class and slot labels are useful for next word prediction in online automatic speech recognition (ASR). Therefore, we propose to perform intent detection, slot filling, and language modeling jointly in a conditional RNN model. The proposed joint model can be further extended for belief tracking in dialogue systems when considering the dialogue history beyond the current utterance. Moreover, it can be used as the RNN decoder in an end-to-end trainable sequence-to-sequence speech recognition model~\cite{jaitly:15}.
    
    The remainder of the paper is organized as follows. In section 2, we introduce the background on using RNNs for intent detection, slot filling, and language modeling. In section 3, we describe the proposed joint online SLU-LM model and its variations. Section 4 discusses the experiment setup and results on ATIS benchmarking task, using both text and noisy speech inputs. Section 5 gives the conclusion.

\section{Background}
\subsection{Intent Detection}
    Intent detection can be treated as a semantic utterance classification problem, where the input to the classification model is a sequence of words and the output is the speaker intent class. Given an utterance with a sequence of words $\mathbf{w}=(w_1, w_2, ..., w_T)$, the goal of intent detection is to assign an intent class $c$ from a pre-defined finite set of intent classes, such that:
        \begin{equation}
            \hat{c} = \operatorname{\arg\max_c} P(c|\mathbf{w}) \\
        \end{equation}
    % Prior work on intent classification includes using uni-bi-trigrams, together with head word, hypernym, part-of-speech tags, and other hand-coded rules in classifiers like support vector machines~\cite{Haffner:03,Silva:11}. 
    Recent neural network based intent classification models involve using neural bag-of-words (NBoW) or bag-of-n-grams, where words or n-grams are mapped to high dimensional vector space and then combined component-wise by summation or average before being sent to the classifier. More structured neural network approaches for utterance classification include using recursive neural network (RecNN)~\cite{guo:14}, recurrent neural network~\cite{ravuri:15}, and convolutional neural network models~\cite{collobert:08,kim:14}. Comparing to basic NBoW methods, these models can better capture the structural patterns in the word sequence.

\subsection{Slot Filling}
    A major task in spoken language understanding (SLU) is to extract semantic constituents by searching input text to fill in values for predefined slots in a semantic frame~\cite{mesnil:15}, which is often referred to as slot filling. The slot filling task can also be viewed as assigning an appropriate semantic label to each word in the given input text. In the below example from ATIS~\cite{hemphill:90} corpus following the popular in/out/begin (IOB) annotation method, \textit{Seattle} and \textit{San Diego} are the from and to locations respectively according to the slot labels, and \textit{tomorrow} is the departure date. Other words in the example utterance that carry no semantic meaning are assigned ``O'' label. 
    
    \begin{figure}[h]
        \centering
        \includegraphics[width=210pt]{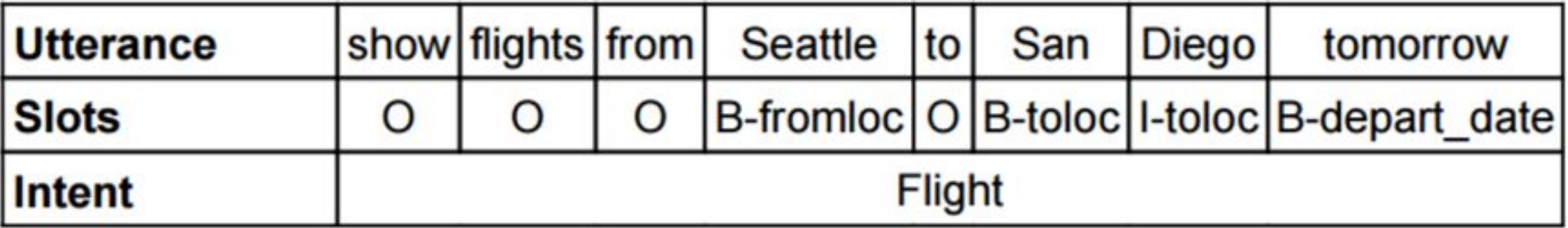}
        \caption{{ATIS corpus sample with intent and slot annotation (IOB format). }}
        \label{fig:slot_filling_example.pdf}
    \end{figure}
    
    Given an utterance consisting of a sequence of words $\mathbf{w}=(w_1, w_2, ..., w_T)$, the goal of slot filling is to find a sequence of semantic labels $\mathbf{s}=(s_1, s_2, ..., s_T)$, one for each word in the utterance, such that:
        \begin{equation}
            \hat{\mathbf{s}} = \operatorname{\arg\max_\mathbf{s}} P(\mathbf{s}|\mathbf{w}) \\
        \end{equation}
    Slot filling is typically treated as a sequence labeling problem. Sequence models including conditional random fields~\cite{raymond:07} and RNN models~\cite{yao:14,mesnil:15,liu:15} are among the most popular methods for sequence labeling tasks. 

\subsection{RNN Language Model}
    A language model assigns a probability to a sequence of words $\mathbf{w}=(w_1, w_2, ..., w_{T})$ following probability distribution. In language modeling, $w_0$ and $w_{T+1}$ are added to the word sequence representing the beginning-of-sentence token and end-of-sentence token. Using the chain rule, the likelihood of a word sequence can be factorized as:
        \begin{equation}
            P(\mathbf{w}) = \prod_{t=1}^{T+1}P(w_{t}|w_0, w_1, ..., w_{t-1}) \\
        \end{equation}
    RNN-based language models~\cite{mikolov:11}, and the variant~\cite{sundermeyer:12} using long short-term memory (LSTM)~\cite{hochreiter:97} have shown superior performance comparing to traditional n-gram based models. In this work, we use an LSTM cell as the basic RNN unit for its stronger capability in capturing long-range dependencies in word sequence. 

\begin{comment}
### Hide this comment part ###
    LSTM model replaces the recurrent module that uses sigmoid or hyperbolic tangent activation function in basic RNN with memory block. Memory block may contain one or more memory cells. The cell state contains summarized information of previous observations, the propagation of which is regulated by cell gates. 
      \begin{equation}
      \begin{split}
      f(t) &= \sigma(U_{f}x(t) + W_{f}h(t-1) + b_{f})\\
      i(t) &= \sigma(U_{i}x(t) + W_{i}h(t-1) + b_{i})\\
      \tilde{C}(t) &= tanh(U_{C}x(t) + W_{C}h(t-1) + b_{C}) \\
      C(t) &= f(t)*C(t-1) + i(t)*\tilde{C}(t) \\
      o(t) &= \sigma(U_{o}x(t) + W_{o}h(t-1) + b_{o}) \\
      h(t) &= o(t)*tanh(C(t))
      \end{split}
      \end{equation}
    where $f$, $i$ and $o$ are the forget gate, input gate and output gate respectively. $C$ is the cell state, and $h$ is the cell output. Figure 2(a) illustrates the architecture of the RNN language model. 
\end{comment}

\subsection{RNN for Intent Detection and Slot Filling}
    \begin{figure}[t]
        \centering
        \includegraphics[width=210pt]{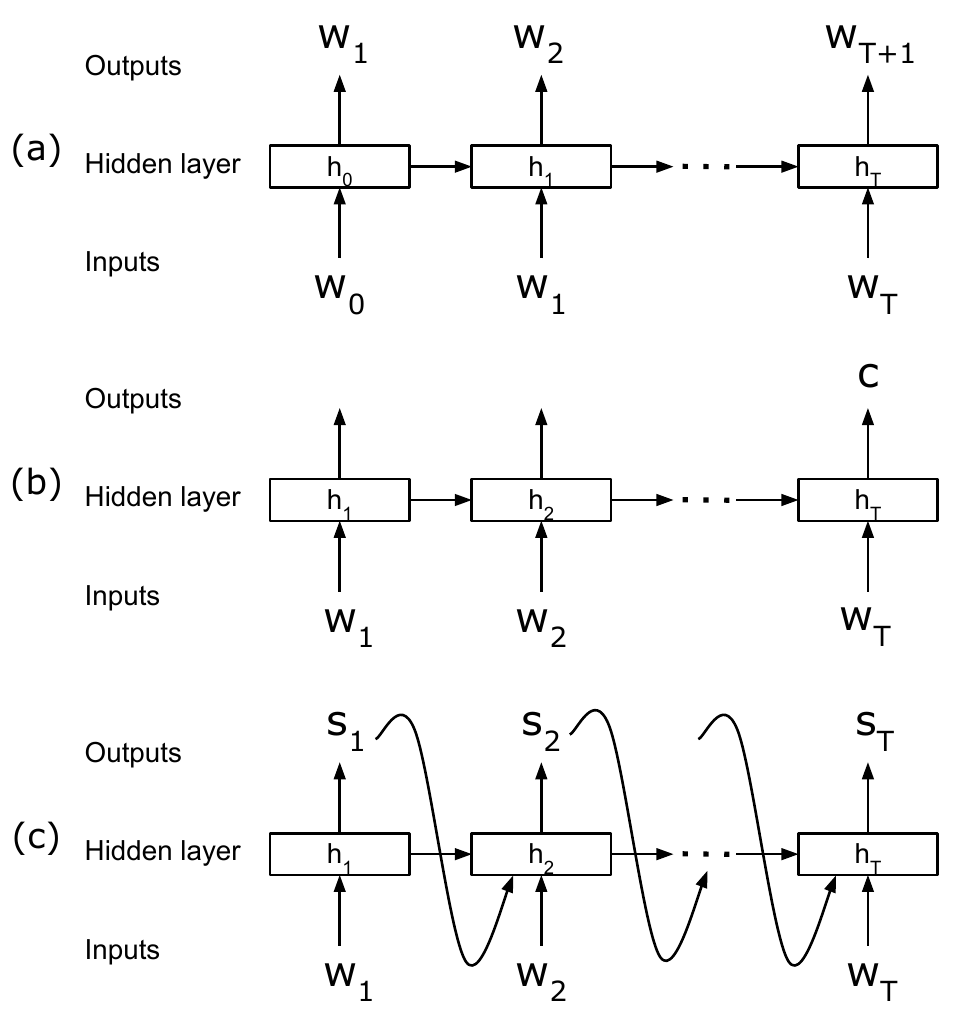}
        \caption{{(a) RNN language model. (b) RNN intent detection model. The RNN output at last step is used to predict the intent class. (c) RNN slot filling model. Slot label dependencies are modeled by feeding the output label of the previous time step to the current step hidden state. }}
        \label{fig:RNN_single_task_LM_Intent_Slot_model}
    \end{figure}
    As illustrated in Figure \ref{fig:RNN_single_task_LM_Intent_Slot_model}(b), RNN intent detection model uses the last RNN output to predict the utterance intent class. This last RNN output can be seen as a representation or embedding of the entire utterance. Alternatively, the utterance embedding can be obtained by taking mean of the RNN outputs over the sequence. This utterance embedding is then used as input to the multinomial logistic regression for the intent class prediction.

    RNN slot filling model takes word as input and the corresponding slot label as output at each time step. The posterior probability for each slot label is calculated using the softmax function over the RNN output. Slot label dependencies can be modeled by feeding the output label from the previous time step to the current step hidden state (Figure \ref{fig:RNN_single_task_LM_Intent_Slot_model}(c)). During model training, true label from previous time step can be fed to current hidden state. During inference, only the predicted label can be used. To bridge the gap between training and inference, scheduled sampling method ~\cite{bengio:15} can be applied. Instead of only using previous true label, using sample from previous predicted label distribution in model training makes the model more robust by forcing it to learn to handle its own prediction mistakes~\cite{liu:15}.

\section{Method}
\begin{figure*}[t]
            \centering
            \includegraphics[width=440pt]{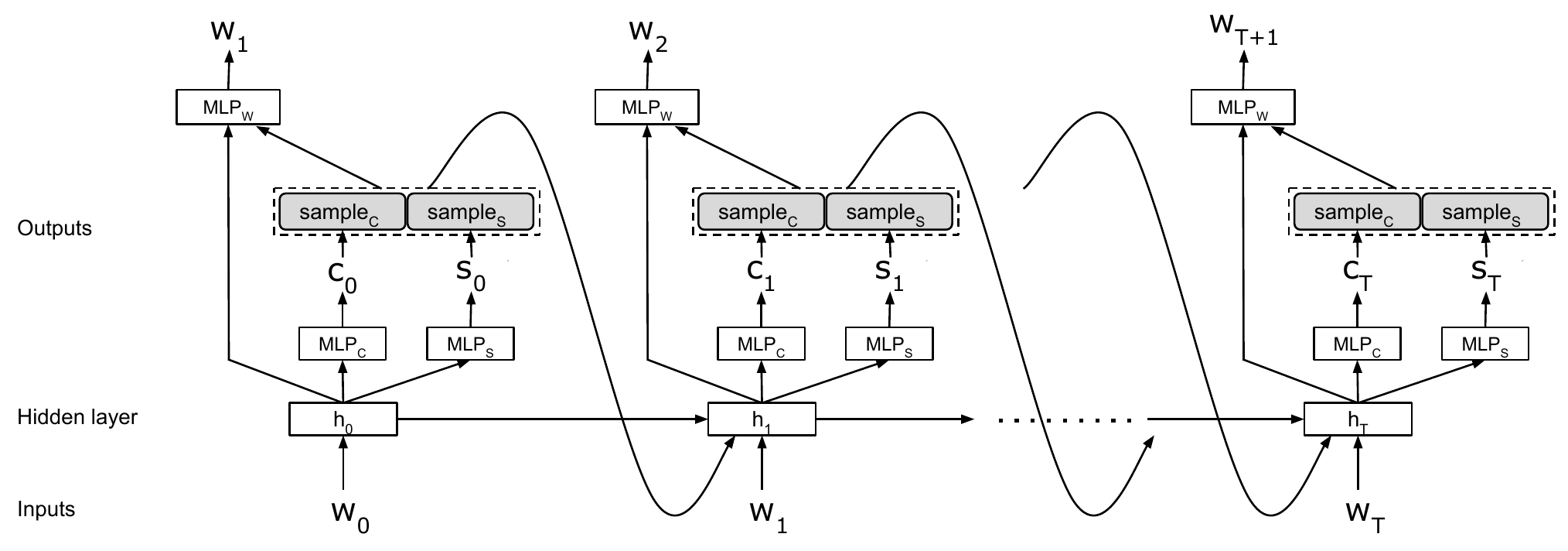}
            \caption{{Proposed joint online RNN model for intent detection, slot filling, and next word prediction. }}
            \label{fig:Joint_SLU_LM_fully_connected}
        \end{figure*}

    In this section we describe the joint SLU-LM model in detail. Figure \ref{fig:Joint_SLU_LM_fully_connected} gives an overview of the proposed architecture.
        \begin{figure*}[t]
            \centering
            \includegraphics[width=440pt]{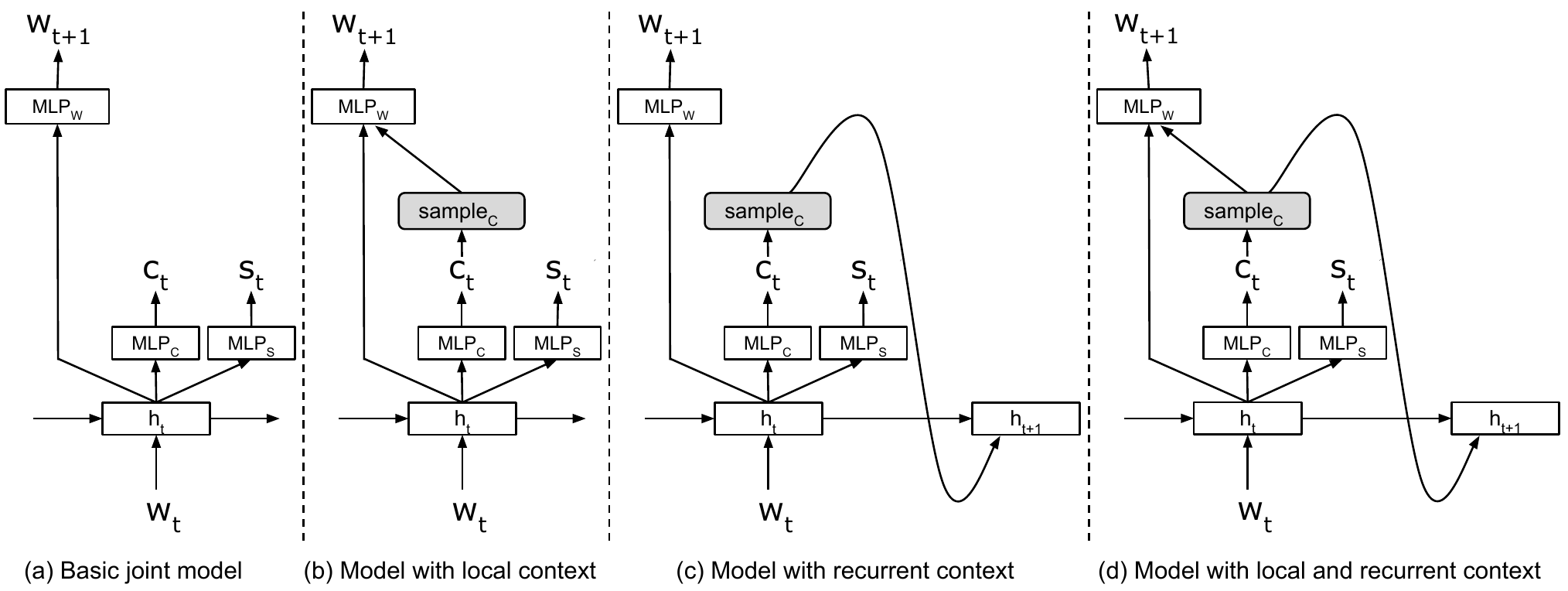}
            \caption{{Joint online SLU-LM model variations. (a) Basic joint model with no conditional dependencies on emitted intent classes and slot labels. (b) Joint model with local intent context. Next word prediction is conditioned on the current step intent class. (c) Joint model with recurrent intent context. The entire intent prediction history and variations are captured in the RNN state. (d) Joint model with both local and recurrent intent context. }}
            \label{fig:Joint_SLU_LM_variations}
        \end{figure*}
    
\subsection{Model}
    Let $\mathbf{w}=(w_0, w_1, w_2, ..., w_{T+1})$ represent the input word sequence, with $w_0$ and $w_{T+1}$ being the beginning-of-sentence ($\langle bos \rangle$) and end-of-sentence ($\langle eos \rangle$) tokens. Let $\mathbf{c}=(c_0, c_1, c_2, ..., c_{T})$ be the sequence of intent class outputs at each time step. Similarly, let $\mathbf{s}=(s_0, s_1, s_2, ..., s_{T})$ be the slot label sequence, where $s_0$ is a padded slot label that maps to the beginning-of-sentence token $\langle bos \rangle$.
    
    Referring to the joint SLU-LM model shown in Figure \ref{fig:Joint_SLU_LM_fully_connected}, for the intent model, instead of predicting the intent only after seeing the entire utterance as in the independent training intent model (Figure \ref{fig:RNN_single_task_LM_Intent_Slot_model}(b)), in the joint model we output intent at each time step as input word sequence arrives. The intent generated at the last step is used as the final utterance intent prediction. The intent output from each time step is fed back to the RNN state, and thus the entire intent output history are modeled and can be used as context to other tasks. It is not hard to see that during inference, intent classes that are predicted during the first few time steps are of lower confidence due to the limited information available. We describe the techniques that can be used to ameliorate this effect in section 3.3 below. For the intent model, with both intent and slot label connections to the RNN state, we have:
        \begin{equation}
            P(c_{T}|\mathbf{w}) = P(c_T|w_{\leq T}, c_{<T}, s_{<T}) \\
        \end{equation}
    For the slot filling model, at each step $t$ along the input word sequence, we want to model the slot label output $s_t$ as a conditional distribution over the previous intents $c_{<t}$, previous slot labels $s_{<t}$, and the input word sequence up to step $t$. Using the chain rule, we have:
        \begin{equation}
            P(\mathbf{s}|\mathbf{w}) = P(s_0|w_0)\prod_{t=1}^{T} P(s_t|w_{\leq t}, c_{<t}, s_{<t}) \\
        \end{equation}
    For the language model, the next word is modeled as a conditional distribution over the word sequence together with intent and slot label sequence up to current time step. The intent and slot label outputs at current step, together with the intent and slot label history that is encoded in the RNN state, serve as context to the language model. 
        \begin{equation}
            P(\mathbf{w}) = \prod_{t=0}^{T} P(w_{t+1}|w_{\leq t}, c_{\leq t}, s_{\leq t}) \\
        \end{equation}
\subsection{Next Step Prediction}
    Following the model architecture in Figure \ref{fig:Joint_SLU_LM_fully_connected}, at time step $t$, input to the system is the word at index $t$ of the utterance, and outputs are the intent class, the slot label, and the next word prediction. The RNN state $h_t$ encodes the information of all the words, intents, and slot labels seen previously. The neural network model computes the outputs through the following sequence of steps:
        \begin{align}
            &h_t = \operatorname{LSTM}(h_{t-1}, [w_t, c_{t-1}, s_{t-1}]) \\
            &P(c_t|w_{\leq t}, c_{<t}, s_{<t}) = \operatorname{IntentDist}(h_{t}) \\
            &P(s_t|w_{\leq t}, c_{<t}, s_{<t}) = \operatorname{SlotLabelDist}(h_{t}) \\
            &P(w_{t+1}|w_{\leq t}, c_{\leq t}, s_{\leq t}) = \operatorname{WordDist}(h_{t}, c_{t}, s_{t})
        \end{align}
    where $\operatorname{LSTM}$ is the recurrent neural network function that computes the hidden state $h_t$ at a step using the previous hidden state $h_{t-1}$, the embeddings of the previous intent output $c_{t-1}$ and slot label output $s_{t-1}$, and the embedding of current input word $w_t$. $\operatorname{IntentDist}$, $\operatorname{SlotLabelDist}$, and $\operatorname{WordDist}$ are multilayer perceptrons (MLPs) with softmax outputs over intents, slot labels, and words respectively. Each of these three MLPs has its own set of parameters. The intent and slot label distributions are generated by the MLPs with input being the RNN cell output. The next word distribution is produced by conditioning on current step RNN cell output together with the embeddings of the sampled intent and  sampled slot label.

\subsection{Training}
    The network is trained to find the parameters $\theta$ that minimise the cross-entropy of the predicted and true distributions for intent class, slot label, and next word jointly. The objective function also includes an $L2$ regularization term $R(\theta)$ over the weights and biases of the three MLPs. This equalizes to finding the parameters $\theta$ that maximize the below objective function:
        \begin{equation}
            \begin{split}
            \max_{\theta} \sum_{t=0}^{T} \Big[ &\alpha _c \log P(c^*|w_{\leq t}, c_{<t}, s_{<t}; \theta) \\
            + &\alpha _s \log P(s^*_{t}|w_{\leq t}, c_{<t}, s_{<t}; \theta) \\
            + &\alpha _w \log P(w_{t+1}|w_{\leq t}, c_{\leq t}, s_{\leq t}; \theta) \Big] \\
            - & \lambda R(\theta)
            \end{split}
        \end{equation}
    where $c^*$ is the true intent class and and $s^*_{t}$ is the true slot label at time step $t$. $\alpha _c$, $\alpha _s$, and $\alpha _w$ are the linear interpolation weights for the true intent, slot label, and next word probabilities. During model training, $c_t$ can either be the true intent or mixture of true and predicted intent. During inference, however, only predicted intent can be used. Confidence of the predicted intent during the first few time steps is likely to be low due to the limited information available, and the confidence level is likely to increase with the newly arriving words. Conditioning on incorrect intent for next word prediction is not desirable. To mitigate this effect, we propose to use a \textit{schedule} to increase the intent contribution to the context vector along the growing input word sequence. Specifically, during the first $k$ time steps, we disable the intent context completely by setting the values in the intent vector to zeros. From step $k+1$ till the last step of the input word sequence, we gradually increase the intent context by applying a linearly growing scaling factor $\eta$ from 0 to 1 to the intent vector. This scheduled approach is illustrated in Figure 5.
        \begin{figure}[h]
            \centering
            \includegraphics[width=180pt]{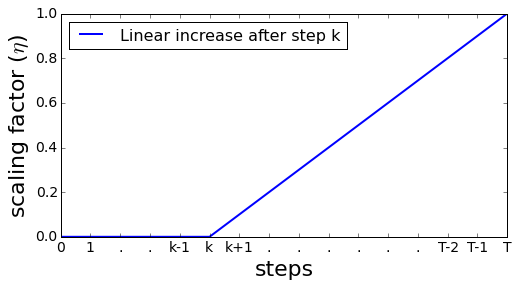}
            \caption{{Schedule of increasing intent contribution to the context vector along with the growing input sequence.}}
            \label{fig:intent_scaling}
        \end{figure}    

\subsection{Inference}
    For online inference, we simply take the greedy path of our conditional model without doing search. The model emits best intent class and slot label at each time step conditioning on all previous emitted symbols:
            \begin{align}
                \hat{c}_t &= \operatorname{\arg\max_ {c_t}} P(c_t|w_{\leq t}, \hat{c}_{<t}, \hat{s}_{<t}) \\
                \hat{s}_t &= \operatorname{\arg\max_ {s_t}} P(s_t|w_{\leq t}, \hat{c}_{<t}, \hat{s}_{<t})
            \end{align}
    Many applications can benefit from this greedy inference approach comparing to search based inference methods, especially those running on embedded platforms that without GPUs and with limited computational capacity. Alternatively, one can do left-to-right beam search ~\cite{sutskever:14,chan:15} by maintaining a set of $\beta$ best partial hypotheses at each step. Efficient beam search method for the joint conditional model is left to explore in our future work.

\subsection{Model Variations}
    In additional to the joint RNN model (Figure \ref{fig:Joint_SLU_LM_fully_connected}) described above, we also investigate several joint model variations for a fine-grained study of various impacting factors on the joint SLU-LM model performance. Designs of these model variations are illustrated in Figure \ref{fig:Joint_SLU_LM_variations}. 
    
    Figure \ref{fig:Joint_SLU_LM_variations}(a) shows the design of a basic joint SLU-LM model. At each step $t$, the predictions of intent class, slot label, and next word are based on a shared representation from the LSTM cell output $h_t$, and there is no conditional dependencies on previous intent class and slot label outputs. The single hidden layer MLP for each task introduces additional discriminative power for different tasks that take common shared representation as input. We use this model as the baseline joint model.
    
    The models in Figure \ref{fig:Joint_SLU_LM_variations}(b) to \ref{fig:Joint_SLU_LM_variations}(d) extend the basic joint model by introducing conditional dependencies on intent class outputs. Note that the same type of extensions can be made on slot labels as well. For brevity and space concern, these designs are not added in the figure, but we report their performance in the experiment section.
    
    The model in Figure \ref{fig:Joint_SLU_LM_variations}(b) extends the basic joint model by conditioning the prediction of next word $w_{t+1}$ on the current step intent class $c_t$. The intent class serves as context to the language model task. We refer to this design as model with \textit{local intent context}.
    
    The model in Figure \ref{fig:Joint_SLU_LM_variations}(c) extends the basic joint model by feeding the intent class back to the RNN state. The history and variations of the predicted intent class from each previous step are monitored by the mode with such class output connections to RNN state. The intent, slot label, and next word predictions in the following step are all dependent on this history of intents. We refer to this design as model with \textit{recurrent intent context}.
    
    The model in Figure \ref{fig:Joint_SLU_LM_variations}(d) combines the two types of connections shown in Figure 4(b) and 4(c). At step $t$, in addition to the \textit{recurrent intent context} ($c_{<t}$), the prediction of word $w_{t+1}$ is also conditioned on the \textit{local intent context} from current step intent class $c_t$. We refer to this design as model with \textit{local and recurrent intent context}.

\section{Experiments}

\subsection{Data}
    We used the Airline Travel Information Systems (ATIS) dataset~\cite{hemphill:90} in our experiment. The ATIS dataset contains audio recordings of people making flight reservations, and it is widely used in spoken language understanding research. We followed the same ATIS corpus\footnote{We thank Gokhan Tur and Puyang Xu for sharing the ATIS dataset.} setup used in ~\cite{mesnil:15,xu:13,tur:10}. The training set contains 4978 utterances from ATIS-2 and ATIS-3 corpora, and test set contains 893 utterances from ATIS-3 NOV93 and DEC94 datasets. We evaluated the system performance on slot filling (127 distinct slot labels) using F1 score, and the performance on intent detection (18 different intents) using classification error rate. 
    
    In order to show the robustness of the proposed joint SLU-LM model, we also performed experiments using automatic speech recognition (ASR) outputs. We managed to retrieve 518 (out of the 893 test utterances) utterance audio files from ATIS-3 NOV93 and DEC94 data sets, and use them as the test set in the ASR settings. To provide a more challenging and realistic evaluation, we used the simulated noisy utterances that were generated by artificially mixing clean speech data with noisy backgrounds following the simulation methods described in the third CHiME Speech Separation and Recognition Challenge~\cite{barker:15}. The average signal-to-noise ratio for the simulated noisy utterances is 9.8dB.

\subsection{Training Procedure}    
    We used LSTM cell as the basic RNN unit, following the LSTM design in~\cite{zaremba:14}. The default forget gate bias was set to 1. We used single layer uni-directional LSTM in the proposed joint online SLU-LM model. Deeper models by stacking the LSTM layers are to be explored in future work. Word embeddings of size 300 were randomly initialized and fine-tuned during model training. We conducted mini-batch training (with batch size 16) using Adam optimization method following the suggested parameter setup in~\cite{kingma:14}. Maximum norm for gradient clipping was set to 5. During model training, we applied dropout (dropout rate 0.5) to the non-recurrent connections~\cite{zaremba:14} of RNN and the hidden layers of MLPs, and applied $L2$ regularization ($\lambda = 10^{-4}$) on the parameters of MLPs. 
    
    For the evaluation in ASR settings, we used the acoustic model trained on LibriSpeech dataset~\cite{panayotov:15}, and the language model trained on ATIS training corpus. A 2-gram language model was used during decoding. Different N-best rescoring methods were explored by using a 5-gram language model, the independent training RNN language model, and the joint training RNN language model. The ASR outputs were then sent to the joint SLU-LM model for intent detection and slot filling.
    \begin{table*} [t]
    \vspace{2mm}
    \centerline{
    \begin{tabular}{r l c c c}
    \hline
    & \textbf{Model} & \textbf{Intent Error}  & \textbf{F1 Score}  & \textbf{LM PPL}\\
    \hline  \hline
    1 & RecNN ~\cite{guo:14}  & 4.60  & 93.22 & -\\ 
    2 & RecNN+Viterbi ~\cite{guo:14} & 4.60 & 93.96 & -  \\ 
    \hline  \hline
    3 & Independent training RNN intent model  & 2.13 & - & - \\
    4 & Independent training RNN slot filling model  & - & 94.91 & - \\
    5 & Independent training RNN language model  & - & - & 11.55 \\
    \hline
    6 & Basic joint training model  & 2.02 & 94.15 & 11.33 \\
    \hline
    7 & Joint model with \textit{local} intent context  & 1.90 & 94.22 & 11.27 \\
    8 & Joint model with \textit{recurrent} intent context  & 1.90 & 94.16 & 10.21 \\
    9 & Joint model with \textit{local \& recurrent} intent context  & 1.79 & 94.18 & 10.22 \\
    \hline
    10 & Joint model with \textit{local} slot label context  & 1.79 & 94.14 & 11.14 \\
    11 & Joint model with \textit{recurrent} slot label context  & 1.79 & \textbf{94.64} & 11.19 \\
    12 & Joint model with \textit{local \& recurrent} slot label context  & 1.68 & 94.52 & 11.17 \\
    \hline
    13 & Joint model with \textit{local} intent + slot label context  & 1.90 & 94.13 & 11.22 \\
    14 & Joint model with \textit{recurrent} intent + slot label context  & \textbf{1.57} & 94.47 & \textbf{10.19} \\
    15 & Joint model with \textit{local \& recurrent} intent + slot label context  & 1.68 & 94.45 & 10.28 \\
    \hline
    \end{tabular}}
    \caption{\label{table1} {ATIS Test set results on intent detection error, slot filling F1 score, and language modeling perplexity. Related \textit{joint} models: \textbf{RecNN}: Joint intent detection and slot filling model using recursive neural network~\cite{guo:14}. \textbf{RecNN+Viterbi}: Joint intent detection and slot filling model using recursive neural network with Viterbi sequence optimization for slot filling~\cite{guo:14}.}}
    \end{table*}    
\subsection{Results and Discussions}
    \subsubsection{Results with True Text Input}
    Table \ref{table1} summarizes the experiment results of the joint SLU-LM model and its variations using ATIS text corpus as input. Row 3 to row 5 are the independent training model results on intent detection, slot filling, and language modeling. Row 6 gives the results of the basic joint SLU-LM model (Figure \ref{fig:Joint_SLU_LM_variations}(a)). The basic joint model uses a shared representation for all the three tasks. It gives slightly better performance on intent detection and next word prediction, with some degradation on slot filling F1 score. If the RNN output $h_t$ is connected to each task output directly via linear projection without using MLP, performance drops for intent classification and slot filling. Thus, we believe the extra discriminative power introduced by the additional model parameters and non-linearity from MLP is useful for the joint model. 
    Row 7 to row 9 of Table \ref{table1} illustrate the performance of the joint models with local, recurrent, and local plus recurrent intent context, which correspond to model structures described in Figure \ref{fig:Joint_SLU_LM_variations}(b) to \ref{fig:Joint_SLU_LM_variations}(d). It is evident that the recurrent intent context helps the next word prediction, reducing the language model perplexity by 9.4\% from 11.27 to 10.21. The contribution of local intent context to next word prediction is limited. We believe the advantageous performance of using recurrent context is a result of modeling predicted intent history and intent variations along with the growing word sequence. For intent classification and slot filling, performance of these models with intent context is similar to that of the basic joint model. 
    
    Row 10 to row 12 of Table \ref{table1} illustrate the performance of the joint model with local, recurrent, and local plus recurrent slot label context. Comparing to the basic joint model, the introduced slot label context (both local and recurrent) leads to a better language modeling performance, but the contribution is not as significant as that from the recurrent intent context. Moreover, the slot label context reduces the intent classification error from 2.02 to 1.68, a 16.8\% relative error reduction. From the slot filling F1 scores in row 10 and row 11, it is clear that modeling the slot label dependencies by connecting slot label output to the recurrent state is very useful.
    \begin{table*} [t]
    \vspace{2mm}
    \centerline{
    \begin{tabular}{l c c c}
    \hline
    \textbf{ASR Model (with LibriSpeech AM)} & \textbf{WER} & \textbf{Intent Error}  & \textbf{F1 Score}  \\
    \hline
    2-gram LM decoding  & 14.51 & 4.63 & 84.46 \\
    2-gram LM decoding + 5-gram LM rescoring  & 13.66 & 5.02 &  85.08 \\
    2-gram LM decoding + Independent training RNN LM rescoring  & 12.95 & 4.63 & 85.43 \\
    2-gram LM decoding + Joint training RNN LM rescoring  & \textbf{12.59} & \textbf{4.44} & \textbf{86.87} \\
    \hline
    \end{tabular}}
    \caption{\label{table2} {ATIS test set results on ASR word error rate, intent detection error, and slot filling F1 score with noisy speech input. }}
    \end{table*}    
    
    Row 13 to row 15 of Table \ref{table1} give the performance of the joint model with both intent and slot label context. Row 15 refers to the model described in Figure \ref{fig:Joint_SLU_LM_fully_connected}. As can be seen from the results, the joint model that utilizes two types of recurrent context maintains the benefits of both, namely, the benefit of applying recurrent intent context to language modeling, and the benefit of applying recurrent slot label context to slot filling. Another observation is that once recurrent context is applied, the benefit of adding local context for next word prediction is limited. It might hint that the most useful information for the next word prediction can be well captured in the RNN state, and thus adding explicit dependencies on local intent class and slot label is not very helpful.

        \begin{figure}[h]
            \centering
            \includegraphics[width=200pt]{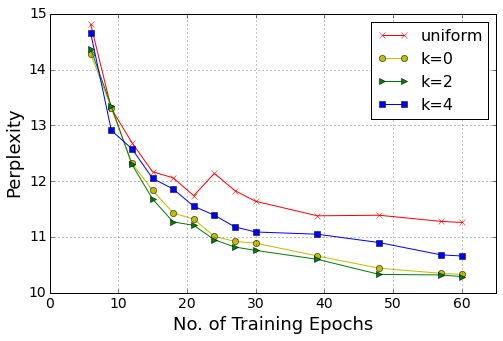}
            \caption{{LM perplexity of the joint SLU-LM models with different schedules in adjusting the intent contribution to the context vector. }}
            \label{fig:intent_scaling_experiments}
        \end{figure}
        
    During the joint model training and inference, we used a schedule to adjust the intent contribution to the context vector by linearly scaling the intent vector with the growing input word sequence after step $k$. We found this technique to be critical in achieving advantageous language modeling performance. Figure 6 shows test set perplexities along the training epochs for models using different $k$ values, comparing to the model with uniform ($\eta$ = 1) intent contribution. With uniform intent contribution across time, the context vector does not bring benefit to the next word prediction, and the language modeling perplexity is similar to that of the basic joint model. By applying the adjusted intent scale ($k=2$), the perplexity drops from 11.26 (with uniform intent contribution) to 10.29, an 8.6\% relative reduction.
 
    \subsubsection{Results in ASR Settings}
    To further evaluate the robustness of the proposed joint SLU-LM model, we experimented with noisy speech input and performed SLU on the rescored ASR outputs. Model performance is evaluated in terms of ASR word error rate (WER), intent classification error, and slot filling F1 score. As shown in Table \ref{table2}, the model with joint training RNN LM rescoring outperforms the models using 5-gram LM rescoring and independent training RNN LM rescoring on all the three evaluation metrics. Using the rescored ASR outputs (12.59\% WER) as input to the joint training SLU model, the intent classification error increased by 2.87\%, and slot filling F1 score dropped by 7.77\% comparing to the setup using true text input. The performance degradation is expected as we used a more challenging and realistic setup with noisy speech input. These results in Table \ref{table2} show that our joint training model outperforms the independent training model consistently on ASR and SLU tasks.
    
\section{Conclusion}
    In this paper, we propose a conditional RNN model that can be used to jointly perform online spoken language understanding and language modeling. We show that by continuously modeling intent variation and slot label dependencies along with the arrival of new words, the joint training model achieves advantageous performance in intent detection and language modeling with slight degradation on slot filling comparing to the independent training models. On the ATIS benchmarking data set, our joint model produces 11.8\% relative reduction on LM perplexity, and 22.3\% relative reduction on intent detection error when using true text as input. The joint model also shows consistent performance gain over the independent training models in the more challenging and realistic setup using noisy speech input. Code to reproduce our experiments is available at: http://speech.sv.cmu.edu/software.html
  
\bibliography{acl2016}
\bibliographystyle{acl2016}
\end{document}